\documentclass[lettersize,journal]{IEEEtran}
\usepackage{amsmath,amsfonts}
\usepackage{algorithm}
\usepackage{array}
\usepackage{textcomp}
\usepackage{stfloats}
\usepackage{url}
\usepackage{verbatim}
\usepackage{graphicx}
\usepackage{cite}
\usepackage{amssymb}

\usepackage{bm}
\usepackage{mathrsfs}

\usepackage{algpseudocode}

\usepackage{tabularx}
\usepackage{booktabs}
\usepackage{multirow}

\usepackage{lipsum}
\usepackage{float}

\usepackage{adjustbox}

\usepackage[table]{xcolor}
\usepackage[hidelinks]{hyperref}

\begin{document}

\title{Dual-Path Stable Soft Prompt Generation for Domain Generalization}

\author{Yuedi Zhang, Shuanghao Bai, Wanqi Zhou, Zhirong Luan, Badong Chen 
\thanks{Corresponding author: Badong Chen.}
\thanks{Yuedi Zhang, Shuanghao Bai, Wanqi Zhou, and Badong Chen are with the Institute of Artificial Intelligence and Robotics, Xi’an Jiaotong University, Xi'an, China (email: zyd993@stu.xjtu.edu.cn; baishuanghao@stu.xjtu.edu.cn; zwq785915792@stu.xjtu.edu.cn; chenbd@mail.xjtu.edu.cn).}
\thanks{Zhirong Luan is with the School of Electrical Engineering, Xi’an University of Technology, Xi'an, China (email: luanzhirong@xaut.edu.cn).}}

\maketitle

\begin{abstract}
Domain generalization (DG) aims to learn a model using data from one or multiple related but distinct source domains that can generalize well to unseen out-of-distribution target domains. 
Inspired by the success of large pre-trained vision-language models (VLMs), prompt tuning has emerged as an effective generalization strategy. However, it often struggles to capture domain-specific features due to its reliance on manually or fixed prompt inputs. 
Recently, some prompt generation methods have addressed this limitation by dynamically generating instance-specific and domain-specific prompts for each input, enriching domain information and demonstrating potential for enhanced generalization. 
Through further investigation, we identify a notable issue in existing prompt generation methods: the same input often yields significantly different and suboptimal prompts across different random seeds, a phenomenon we term Prompt Variability. 
To address this, we introduce negative learning into the prompt generation process and propose Dual-Path Stable Soft Prompt Generation (DPSPG), a transformer-based framework designed to improve both the stability and generalization of prompts. 
Specifically, DPSPG incorporates a complementary prompt generator to produce negative prompts, thereby reducing the risk of introducing misleading information. 
Both theoretical and empirical analyses demonstrate that negative learning leads to more robust and effective prompts by increasing the effective margin and reducing the upper bound of the gradient norm.
Extensive experiments on five DG benchmark datasets show that DPSPG consistently outperforms state-of-the-art methods while maintaining prompt stability.
The code is available at \url{https://github.com/renytek13/Dual-Path-Stable-Soft-Prompt-Generation}.
\end{abstract}


\section{Introduction}
\label{sec:intro}

\IEEEPARstart{D}{omain} generalization (DG) aims to train models on data from one or more related yet distinct source domains, enabling them to generalize effectively to unseen out-of-distribution (OOD) target domains. The core objective is to learn transferable, domain-invariant representations that remain robust under distributional shifts~\cite{zhou2022domain, wang2022generalizing}. Traditional approaches address this challenge by enhancing data diversity through data augmentation~\cite{volpi2018generalizing, li2021simple} and data extension~\cite{zhang2017mixup}, learning domain-invariant representations~\cite{muandet2013domain, chen2024domain}, or employing strategies such as ensemble learning~\cite{arpit2022ensemble} and meta-learning~\cite{zhou2022domain} to promote generalization across domains.

While traditional DG methods typically rely solely on visual information, they often overlook the rich semantic information. This limitation is critical, as recent studies have shown that preserving semantic integrity while mitigating distribution shifts can significantly improve generalization~\cite{ge2023domain, bai2024prompt}.
To bridge this gap, large pre-trained vision-language models (VLMs), such as CLIP~\cite{radford2021learning} and ALIGN~\cite{jia2021scaling}, leverage large-scale image-text pairs during training and have demonstrated strong zero-shot generalization across a wide range of downstream tasks~\cite{zhou2022learning, ge2023domain, zhou2024revisiting}.
However, their limited adaptability to downstream tasks in DG settings constrains their effectiveness and warrants further exploration.

To better adapt vision-language models to DG tasks, prompt tuning has emerged as a promising direction and can be broadly categorized into two paradigms.
The first is fixed prompt learning~\cite{bai2024prompt}, where a small set of learnable prompt vectors is prepended to the input and optimized during training. These prompts are then directly reused for all inputs during inference. Representative methods include CoOp~\cite{zhou2022domain} and MaPLe~\cite{khattak2023maple}. Although efficient and lightweight, fixed prompts lack flexibility in capturing domain-specific and instance-specific information.
The second paradigm is dynamic prompt learning, where prompts are generated based on the input or domain context. For example, DPL~\cite{zhang2023domain} employs a domain-wise prompt generator conditioned on domain labels, while SPG~\cite{bai2024soft} uses a generative network to produce instance-specific prompts. By tailoring prompts to each domain or instance, dynamic approaches better capture domain-specific semantics and enhance generalization across domains.

\begin{figure}[tbp]
  \centering
  \includegraphics[width=0.50\textwidth]{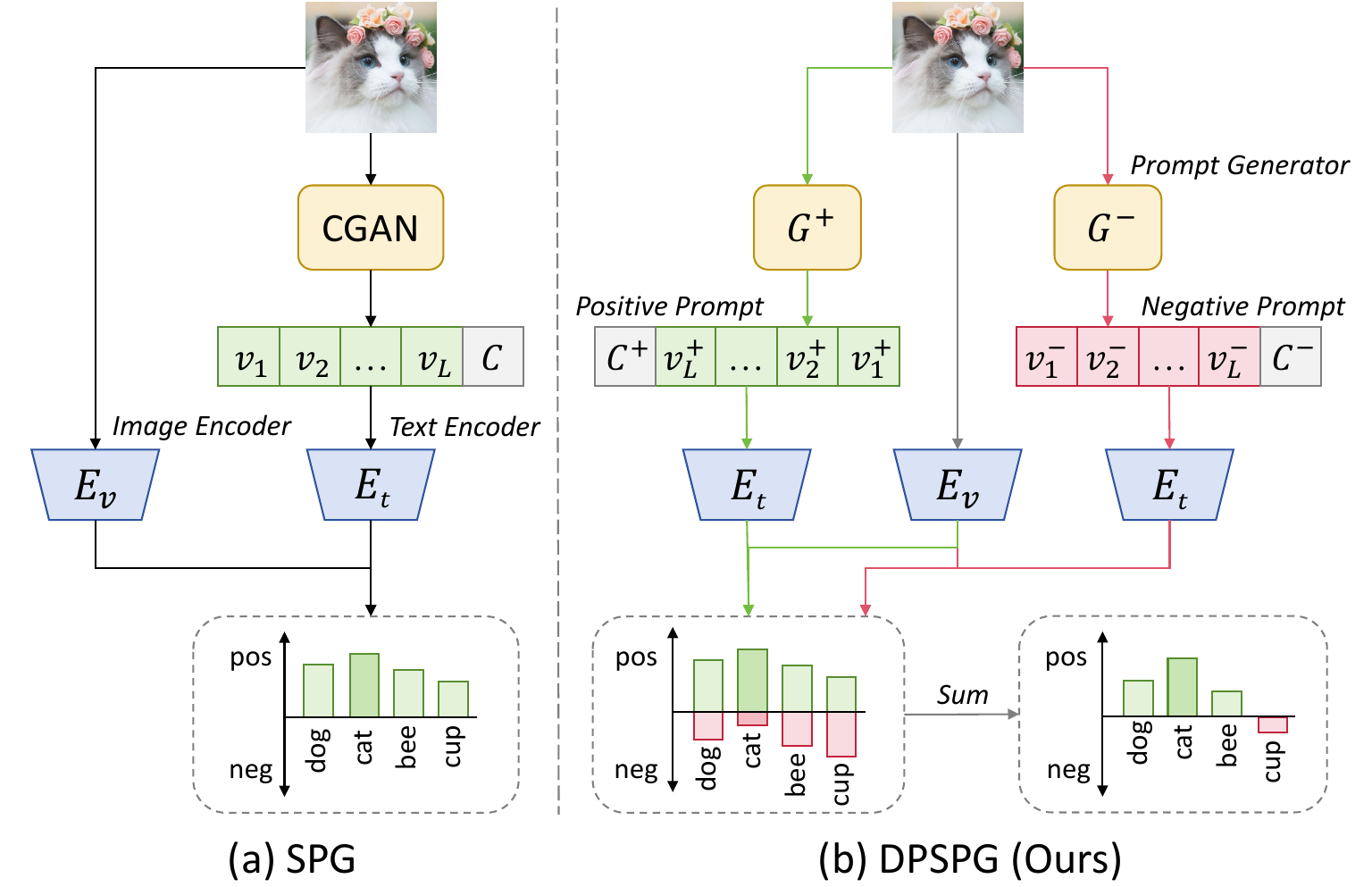}
  \caption{Comparison of the inference stage between our proposed DPSPG and SPG~\cite{bai2024soft}. The dual-path strategy in DPSPG enhances the robustness and stability of prompt generation while maintaining domain-specific semantic coherence.}
  \label{fig:intro}
\end{figure}

\begin{figure}[tbp]
  \centering
  \includegraphics[width=\linewidth]{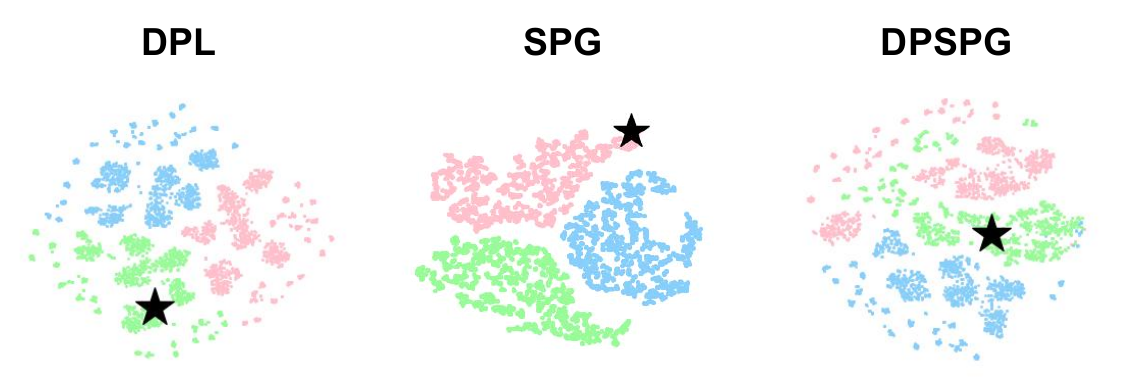}
  \caption{Comparison of prompt generation quality between our proposed DPSPG and existing methods, DPL~\cite{zhang2023domain} and SPG~\cite{bai2024soft}. Colored clusters represent the distribution of prompts generated for the photo domain in the PACS test set under different random seeds, while the pentagram denotes the optimal prompt. DPSPG generates prompts that are more consistently clustered around the optimal point, indicating higher generation quality and stronger domain focus.}
  \label{fig:prompt_visual}
\end{figure}

To further explore the limitations of dynamic prompt learning, we examine the prompt quality of DPL~\cite{zhang2023domain} and SPG~\cite{bai2024soft}. 
As shown in Figure~\ref{fig:prompt_visual}, prompts generated by these methods fail to cluster around the optimal prompt, which is defined as the one learned directly from the test domain and achieving the highest accuracy. This observation suggests that the generated prompts are suboptimal in capturing domain-specific information.
Table~\ref{tab:prompt_distance} further confirms this observation, where both methods yield higher intra-to-inter domain distance ratios, reflecting unstable generation across different random seeds.
We hypothesize that this instability arises from two sources. For DPL, the averaging of inputs from multiple domains during training may obscure domain-specific signals, leading to inconsistent prompt representations. For SPG, the inherent noise introduced by the generative prompt network may degrade prompt quality and stability.
We collectively refer to these issues as Prompt Variability.

To address the Prompt Variability problem, we introduce negative learning into prompt generation for DG and propose a novel framework, Dual-Path Stable Soft Prompt Generation (DPSPG).
During training, we first generate domain-specific positive and negative optimal prompts, which serve as domain prompt labels. Two separate transformer-based prompt generators are then trained to align their outputs with the corresponding positive and negative labels, respectively.
At inference time, as illustrated in Figure~\ref{fig:intro}, the trained generators produce both positive and negative prompts for each image in the target domain. This dual-path strategy enhances robustness and stability in prompt generation, while maintaining domain semantic coherence and improving transferability.

Empirically, DPSPG improves both the stability of the prompt generator training process and the generalization capability of the learned prompts.
As shown in Figure~\ref{fig:prompt_visual}, the prompts generated by DPSPG consistently cluster around the optimal prompt, indicating stronger alignment with domain semantics.
Table~\ref{tab:prompt_distance} further shows that DPSPG achieves a significantly lower intra-to-inter domain distance ratio compared to existing methods such as DPL and SPG, demonstrating enhanced intra-domain consistency and improved cross-domain discriminability.
Moreover, as reported in Table~\ref{tab:efficiency}, DPSPG achieves a lower standard deviation in accuracy over the last 10 epochs, suggesting a more stable optimization process during training compared to previous dynamic prompt learning approaches.

Motivated by these empirical findings, we further provide a theoretical analysis to substantiate the effectiveness of DPSPG introduced through negative learning.
Specifically, we show that the incorporation of negative prompts enlarges the effective decision margin between the ground-truth class and competing classes, thereby enhancing class separability.
The increased margin exponentially tightens the upper bound of the gradient norm, leading to a smoother and more stable optimization process.
Moreover, the corresponding reduction in the Jacobian norm with respect to input perturbations improves robustness against noise and adversarial variations, further enhancing the reliability and generalization of the learned prompts.

\begin{table}[t]
  \caption{QUANTITATIVE ANALYSIS OF PROMPT GENERATION FOR DPL, SPG, AND OUR PROPOSED DPSPG ON THE PHOTO DOMAIN OF THE PACS DATASET. HERE, $R$ DENOTES INTRA-DOMAIN DISTANCE, $D$ DENOTES INTER-DOMAIN DISTANCE. A LOWER $\lambda$ INDICATES MORE STABLE PROMPT GENERATION WITHIN DOMAINS AND BETTER SEPARATION BETWEEN DOMAINS. DPSPG ACHIEVES MORE STABLE AND DISCRIMINATIVE PROMPT GENERATION, RESULTING IN OPTIMAL ACCURACY}
  \label{tab:prompt_distance}
  \centering
  \renewcommand{\arraystretch}{1}
  \begin{tabular}{lccc}
    \toprule
    Method & DPL & SPG & DPSPG  \\ 
    \midrule
    $R$ & 1.9082 & 0.5763 & \textbf{0.4762} \\
    $\lambda=R / D$ & 0.9722 & 0.2936 & \textbf{0.2426} \\
    $acc$ & 99.03 & 99.47 & \textbf{99.70} \\
    \bottomrule
  \end{tabular}
\end{table}

Our main contributions are as follows.
First, we identify and formalize the Prompt Variability problem in dynamic prompt learning. To address this problem, we propose a novel framework, Dual-Path Stable Soft Prompt Generation (DPSPG), which introduces negative learning into prompt generation. This dual-path design leads to more stable and transferable prompts across domains.
Second, we provide a theoretical analysis demonstrating that negative learning increases the effective decision margin, tightens the upper bound of the gradient norm, and reduces sensitivity to input perturbations, thereby offering a principled explanation for improved generalization.
Finally, extensive experiments on five domain generalization benchmarks show that DPSPG consistently outperforms existing methods, achieving state-of-the-art performance in both accuracy and prompt stability.

\section{Related Work}
\label{sec:rw}

\subsection{Domain Generalization}
Domain Generalization (DG) aims to train models that generalize well to unseen domains by leveraging data from source domains with varying distributions~\cite{zhou2022domain}. 
Existing DG approaches can be broadly grouped into three major categories~\cite{wang2022generalizing}.
One line of research focuses on data-level techniques, such as data augmentation~\cite{volpi2018generalizing, jiang2024cbda} and data extension~\cite{zhang2017mixup, cao2024mixup}, which increase training data diversity to improve model robustness against domain shifts.
Another prominent category emphasizes representation learning. These methods aim to extract domain-invariant features via domain alignment~\cite{li2018domain, hu2024domain}, adversarial learning~\cite{ganin2016domain, li2018deep}, invariant risk minimization~\cite{arjovsky2019invariant, li2022invariant}, or feature disentanglement~\cite{li2021simple, bui2021exploiting}, enabling better transferability across domains.
A third direction investigates learning strategies, including ensemble learning~\cite{arpit2022ensemble, li2022simple, zhou2024mixstyle}, knowledge distillation~\cite{niu2023knowledge, zhang2023boosting}, and meta-learning~\cite{li2018learning, chen2022discriminative}.
Recently, vision-language models such as CLIP~\cite{radford2021learning} have demonstrated strong zero-shot generalization capabilities for DG and effectively bridge the semantic gap.
To further adapt these models to DG tasks, prompt learning has emerged as a promising fine-tuning strategy.
Building on this line of work, we aim to address key limitations of existing prompt learning methods for DG, including the lack of domain-specific knowledge in generated prompts and the instability and suboptimality of the prompt generation process.

\subsection{Prompt Learning in Vision Language Models for DG}
Given the large scale of Vision Language Models (VLMs), recent research has focused on lightweight and efficient fine-tuning methods for adapting to downstream tasks, primarily categorized into prompt learning~\cite{li2022learning, zhang2024promptta, zhang2025consistent} and adapter tuning~\cite{zhang2022tip, gao2024clip, yu2024clipceil}. 
Prompt learning methods can be broadly categorized into two main types.
Fixed prompt learning~\cite{zhou2022learning, bose2024stylip, cheng2024disentangled} optimizes a small set of continuous context vectors during training, which remain static and are applied uniformly to all test inputs during inference.
For example, CoOp~\cite{zhou2022learning} introduces learnable text-based soft prompts, while MaPLe~\cite{khattak2023maple} further enhances prompt learning by incorporating both vision and language prompts and exploiting their synergy to refine representations.
DDSPL~\cite{xu2024ensembling} further proposes to ensemble domain-specific prompts through weighted aggregation.
Moreover, dynamic prompt learning generates instance-specific prompts conditioned on domain or input information.
CoCoOp~\cite{zhou2022conditional} extends CoOp by introducing an MLP conditioned on image features to generate instance-specific prompts.
DPL~\cite{zhang2023domain} employs an MLP to generate prompts based on the average feature representation of each randomly sampled batch.
SPG~\cite{bai2024soft} further utilizes generative adversarial networks to synthesize instance-specific prompts enriched with domain-specific information.
Although dynamic prompt learning methods are effective at capturing rich domain-specific information, they suffer from the Prompt Variability problem identified in our study. To address this issue, we introduce negative learning to stabilize the training process and improve both the generalization capability and stability of the learned prompts.

\subsection{Negative Learning}
Negative learning has been widely explored across different tasks to improve robustness, enhance discriminative representations, and mitigate label noise.
In the context of image classification, early work such as NLNL~\cite{kim2019nlnl} proposes training on complementary negative labels to address issues arising from inaccurate and noisy labels.
In self-supervised learning, contrastive frameworks like SimCLR~\cite{chen2020simple} and MoCo~\cite{he2020momentum} leverage large pools of negative samples to sharpen feature boundaries, leading to more discriminative and robust representations.
Similarly, RPL~\cite{chen2020learning} introduces reciprocal points as negative representations corresponding to all existing classes, effectively bounding the open-space risk and improving open-set recognition.
Recently, negative learning has also been extended to vision-language models (VLMs).
ArGue~\cite{tian2024argue} augments CLIP with negative prompts to counteract inherent biases, thereby improving OOD generalization.
Building on this idea, CLIPN~\cite{wang2023clipn} and NegPrompt~\cite{li2024learning} propose generating "what-not" prompts to reinforce OOD recognition by explicitly modeling negative concepts.
In semi-supervised VLM training, DNLL~\cite{xu2023semi} utilizes pseudo-negative labels to filter noisy data, yielding cleaner training signals and improving robustness.
Furthermore, DEFT~\cite{NEURIPS2024_6af08ba9} introduces paired positive and negative prompts to detect and correct noisy labels during VLM fine-tuning, enhancing classification performance under label noise.
Different from previous works, we introduce negative learning into prompt learning to address the problem of prompt variability, aiming to stabilize prompt generation and enhance alignment with domain semantics. We empirically validate the effectiveness of our approach and further provide a theoretical analysis, demonstrating that incorporating negative prompts enlarges the decision margin and tightens the upper bound of the gradient norm.

\section{Method}
\label{sec:method}

This section is organized as follows. Section~\ref{subsec:pre} introduces the necessary preliminaries on DG and CLIP-based prompt learning. Section~\ref{subsec:dpspg} presents our proposed Dual-Path Stable Soft Prompt Generation (DPSPG) framework. Section~\ref{subsec:theo} provides a theoretical analysis demonstrating how negative learning enhances prompt stability and generalization.

\subsection{Preliminaries}
\label{subsec:pre}

\subsubsection{Domain Generalization}

We consider a training dataset $\mathcal{D}_s = \{(x_i, y_i)\}_{i=1}^N$,, where $x_i \in \mathcal{X}_s$ represents an input sample from the source domain $\mathcal{X}_s$, $y_i \in \mathcal{Y}_s$ denotes the corresponding label, and $N$ is the total number of training samples.
The goal of DG is to train a model composed of a feature extractor $g\phi$ and a classifier $h\theta$, where $g\phi: \mathcal{X}_s \rightarrow \mathcal{Z}_s$ maps inputs to a feature space $\mathcal{Z}_s$, and $h\theta: \mathcal{Z}_s \rightarrow \mathcal{Y}_s$ predicts labels based on the extracted features.
The combined model $f(x) = h_\theta(g_\phi(x))$ is expected to generalize well to an unseen target domain $\mathcal{X}_t$, meaning that $f$ should accurately predict $y_t \in \mathcal{Y}_t$ for samples drawn from $\mathcal{X}_t$, despite the distributional shift between $\mathcal{X}_s$ and $\mathcal{X}_t$.
Depending on the number of source domains, DG settings can be categorized into single-source DG and multi-source DG. In this work, we focus on the multi-source DG scenario.
Generally, DG aims to learn domain-invariant representations by minimizing a loss function $\ell$ over multiple source domains, formulated as:
\begin{equation}\begin{aligned}\label{func:dg}
\min_{\phi,\theta} \sum_{s=1}^S \mathbb{E}_{(x_i, y_i) \in \mathcal{D}_s} \left[ \ell(h_\theta(g_\phi(x_i)), y_i) \right],
\end{aligned}\end{equation}
where $S$ denotes the number of source domains.
The key challenge lies in designing appropriate architectures for $g_\phi$ and $h_\theta$, possibly with additional regularization or adversarial strategies, to ensure that the learned representations $z$ generalize effectively to unseen target domains.

\begin{figure*}[!t]
  \centering
  \includegraphics[width=0.78\textwidth]{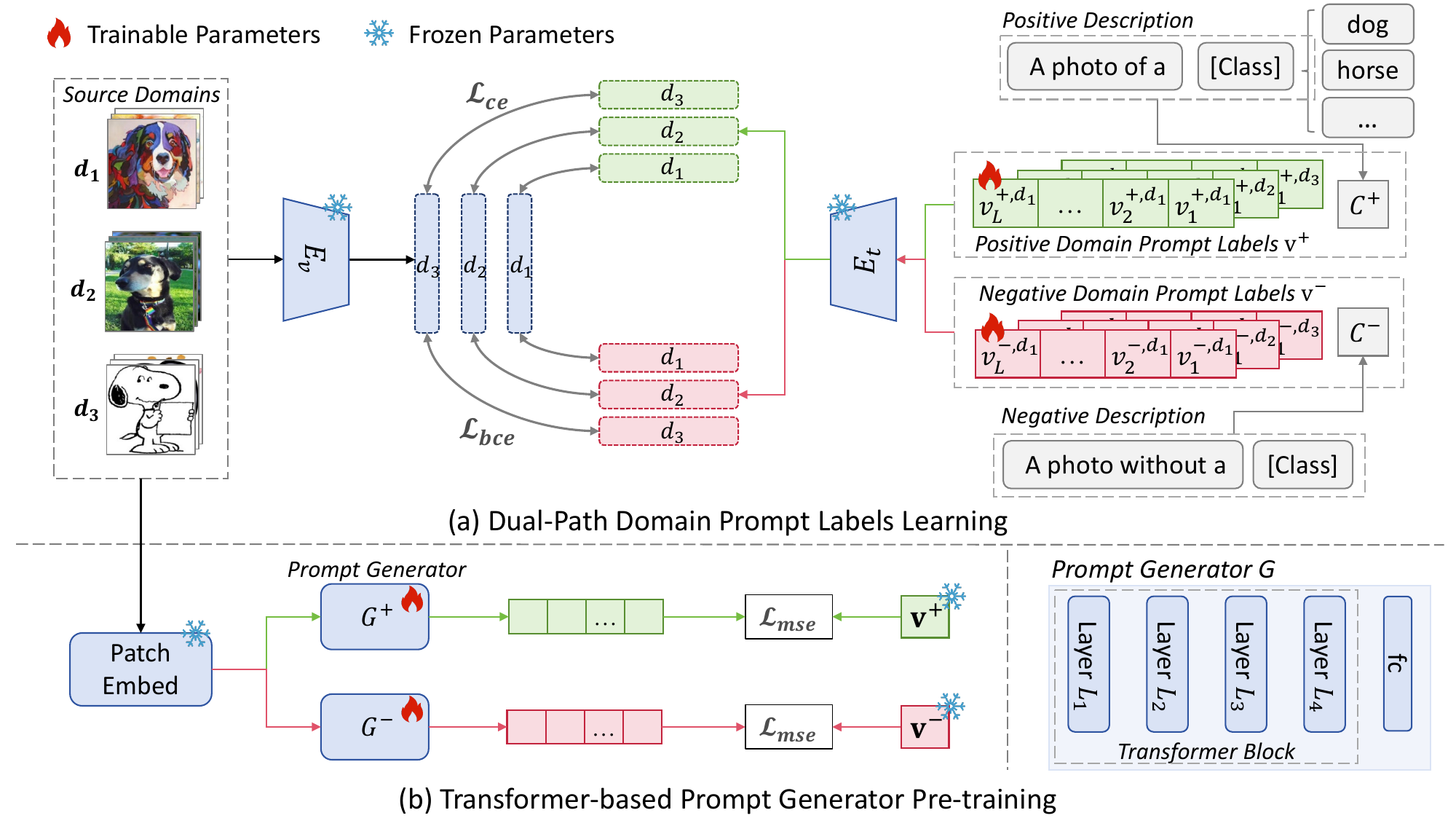}
  \caption{The training process of DPSPG consists of two stages. In the first stage, positive and negative domain prompt labels are learned. In the second stage, positive and negative prompts for images are generated using separate transformer-based prompt generators and are aligned with the corresponding positive and negative prompt labels.}
  \label{fig:model}
\end{figure*}

\subsubsection{Prompt Learning in VLMs}
In image classification, let $D = {(x, y)}$ denote the downstream dataset, where $x$ represents an input image and $y$ is its corresponding class label.
Let $\phi$ and $\psi$ denote the visual encoder and text encoder of CLIP~\cite{radford2021learning}, respectively.
For the \textit{i}-th class, we construct a class-specific text prompt by prepending a manually designed template, ``a photo of a \{$class_i$\}", resulting in the prompt ``a photo of a \{$class_i$\}". 
This textual prompt is then tokenized and embedded as $w_i$, which is passed through the text encoder $\psi$ to obtain the corresponding class feature $t_i = \psi(w_i)$.
The collection of all class-specific text features is denoted as $T = {t_1, t_2, \ldots, t_K}$.
Given an input image $x$, the predicted probability distribution is computed as:
\begin{equation}\begin{aligned}\label{func:zsclip}
p(y \mid \bm{x}) = \frac{\exp( \langle \phi(\bm{x}), t_i\rangle / \tau)}{\sum_{j=1}^K \exp( \langle \phi(\bm{x}), t_j\rangle / \tau)},
\end{aligned}\end{equation}
where $\tau$ denotes temperature parameter, $K$ denotes the number of classes, and $\langle \cdot,\cdot \rangle$ denotes the cosine similarity.

Unlike zero-shot CLIP, which relies on manually designed prompts, CoOp \cite{zhou2022learning} introduces a set of learnable soft prompts $\mathbf{v} = \{\mathbf{v}_1, \mathbf{v}_2, \ldots, \mathbf{v}_M\}$ as continuous context vectors. 
These vectors are concatenated with the word embedding $c_i$ of each class name to form a class-specific prompt $w_i = [\mathbf{v}, c_i]$, which is then fed into the text encoder $\psi$ to obtain the text feature $t_i = \psi([\mathbf{v}, c_i])$, where $t_i \in \{t_1, t_2, \ldots, t_K\}$. 
The class probability is computed following Equation~\ref{func:zsclip}, and the soft prompts $\mathbf{v}$ are optimized via standard cross-entropy loss:
\begin{equation}
\mathcal{L}_{\text{ce}} = -\frac{1}{N} \sum_{i=1}^{N} \sum_{c=1}^{K} y_{i,c} \log p(y = c \mid \mathbf{x}_i).
\end{equation}

\subsection{Dual-Path Stable Soft Prompt Generation}
\label{subsec:dpspg}

Our DPSPG framework adopts a two-stage training process, as illustrated in Figure~\ref{fig:model}.  
In the first stage, we construct positive and negative domain prompt labels as ground truth. 
In the second stage, two transformer-based prompt generators, conditioned on the image embeddings from the source domains, are trained to produce positive and negative prompts that align with the corresponding domain prompt labels, respectively.

\subsubsection{Dual-Path Domain Prompt Labels Learning}

As shown in Figure~\ref{fig:model}~(a), we introduce the concept of dual-path prompt learning with domain-specific positive and negative prompt labels. 
For each source domain, we first construct two learnable prompt vectors using text prompts~\cite{zhou2022learning}: one positive and one negative. 
Specifically, each domain corresponds to a pair of positive domain prompt labels $\mathbf{v}^{+, d_j}$ and negative domain prompt labels $\mathbf{v}^{-, d_j}$, where $d_j$ denotes the \textit{j}-th domain. 
Furthermore, we initialize the positive template as \texttt{`a photo of a \{class\}'}, and the negative template as \texttt{`a photo without a \{class\}'}. 
The positive text feature for the \textit{i}-th class of the \textit{j}-th domain is then represented as ${t}^+_i = \psi([\mathbf{v}^{+, d_j}, v^+, c_i])$, and the negative text feature as ${t}^-_i = \psi([\mathbf{v}^{-, d_j}, v^-, c_i])$, 
where $v^+$ and $v^-$ denote the positive and negative template embeddings (excluding the class name), and $c_i$ denotes the word embedding of the \textit{i}-th class name. 
The pseudo-code framework for dual-path domain prompt label learning is described in Algorithm~\ref{alg:pn_prompt}.

\begin{algorithm}[tbp]
\caption{DPSPG: Dual-Path Domain Prompt Labels Learning}
\label{alg:pn_prompt}
\textbf{Requirement:} Training datasets $\{D_j\}_{j=1}^{N_d}$, text encoder $\psi$ \\
\textbf{Input:} Training iterations $L$, number of categories $K$ \\
\textbf{Output:} Trained pos-neg labels $\mathbf{v}^{+, d_j}$, $\mathbf{v}^{-, d_j}$
\begin{algorithmic}[1]
\For{$j = 1, 2, ..., N_d$}
    \State positive fixed prompt $v^+ \leftarrow [\mathtt{a\ photo\ of\ a}]$
    \State negative fixed prompt $v^- \leftarrow [\mathtt{a\ photo\ without\ a}]$
    \For{$i = 1, 2, ..., K$}
        \State ${t}^+_i \leftarrow \psi([\mathbf{v}^{+, d_j}, v^+, c_i]), \quad {t}^-_i \leftarrow \psi([\mathbf{v}^{-, d_j}, v^-, c_i])$
    \EndFor
    \For{$l = 1, 2, ..., L$}
        \State $\mathcal{L}_{\mathrm{pos}} \leftarrow Equation~\eqref{func:positive_loss}, \quad \mathcal{L}_{\mathrm{neg}} \leftarrow Equation~\eqref{func:negative_loss}$

        \State Update $\mathbf{v}^{+, d_j}$ and $\mathbf{v}^{-, d_j}$ by gradient descent
    \EndFor
\EndFor
\State \textbf{Store} trained optimal pos-neg domain prompt labels $\mathbf{v}^{+, d_j}$, $\mathbf{v}^{-, d_j}$
\end{algorithmic}
\end{algorithm}

First, we train the learnable positive domain prompt labels $\mathbf{v}^{+, d_j}$ using the standard cross-entropy loss as follows:
\begin{equation}
\begin{aligned}
\label{func:positive_loss}
&\mathcal{L}_{\text{ce}}(\mathbf{v}^{+, d_j}) \\
&= - \mathbb{E}_{(\mathbf{x}, y) \sim \mathcal{D}_j} \left[ \sum_{i} y_{d_j,i}^{+} \log p(y_{d_j,i}^{+} \mid \mathbf{x}_i,\mathbf{v}^{+, d_j},c_i) \right],
\end{aligned}
\end{equation}
where $y_{d_j,i}^{+}$ denotes the ground-truth label of the \textit{i}-th class in \textit{j}-th domain.

Then, for the domain negative prompt labels $\mathbf{v}^{-, d_j}$, we use binary cross-entropy loss due to the nature of their encoding. 
Given that $\mathbf{y}_{d_j}$ is a multi-label one-hot encoded vector (e.g., $[1, 0, 1]$), we train the learnable negative domain prompt labels $\mathbf{v}^{-, d_j}$ with the binary cross-entropy (BCE) loss, which can be expressed as:
\begin{equation}
\begin{aligned}
\label{func:negative_loss}
&\mathcal{L}_{\text{bce}}(\mathbf{v}^{-, d_j}) \\
&= -\mathbb{E}_{(x, \mathbf{y}) \sim \mathcal{D}_j} \Bigg[\sum_{i} \Bigg( y_{d_j, i} \log p(y_{d_j, i} \mid \mathbf{x}_i, \mathbf{v}^{-, d_j}, c_i) \quad \\
& + (1 - y_{d_j, i}) \log(1 - p(y_{d_j, i} \mid \mathbf{x}_i, \mathbf{v}^{-, d_j}, c_i)) \Bigg) \Bigg],
\end{aligned}
\end{equation}

where $\mathbf{y}_{d_j} = [y_{d_j,1}, y_{d_j,2}, \dots, y_{d_j,K}]$ denotes a multi-label vector, with each $y_{d_j,i}$ being a binary value (0 or 1). The term $p(y_{d_j,i} \mid \mathbf{x}_i, \mathbf{v}^{-, d_j}, c_i)$ represents the predicted probability that the $i$-th label is true given the input $\mathbf{x}_i$ and the domain-specific negative prompt vector $\mathbf{v}^{-, d_j}$.

\subsubsection{Transformer-based Prompt Generator Pre-training}

As illustrated in Figure~\ref{fig:model} (b), we employ a transformer-based model as our prompt generator, consisting of four transformer layers and a linear layer $fc$, to generate positive and negative soft prompts for image data. 
Specifically, the image embedding $\phi(\bm{x})$ is fed into the transformer encoders $G^+$ and $G^-$ to produce the positive prompt $\hat{\mathbf{v}}^{+,d_i}$ and the negative prompt $\hat{\mathbf{v}}^{-,d_i}$, respectively. 
The pseudo-code framework for transformer-based prompt generator pre-training is provided in Algorithm~\ref{alg:transf_pretr}.

\begin{algorithm}[tbp]
\caption{DPSPG: Transformer-based Prompt Generator Pre-training}
\label{alg:transf_pretr}
\textbf{Requirement:} Trained pos-neg domain prompt labels $\mathbf{v}^{+, d_j}$, $\mathbf{v}^{-, d_j}$, transformer-based prompt generators $G^{+}$, $G^{-}$ \\
\textbf{Input:} Image embeddings $\phi(\bm{x})$, training iterations $L$ \\
\textbf{Output:} Generated pos-neg prompt for target image using trained $G^{+}$, $G^{-}$
\begin{algorithmic}[1]
\For{$l = 1, 2, ..., L$}
    \State $\hat{\mathbf{v}}^{+,d_i} = G^+(\phi(\bm{x})), \quad \hat{\mathbf{v}}^{-,d_i} = G^-(\phi(\bm{x}))$

    \State $\mathcal{L}_{mse} \leftarrow Equation~\eqref{eq:DPSPG}$
    \State Update parameters of $G^+$ and $G^-$ using $\mathcal{L}_{mse}$ by gradient descent
\EndFor
\State \textbf{Generate} pos-neg prompt for each input image in target domain
\end{algorithmic}
\end{algorithm}

The second training stage aims to ensure that the generated positive and negative prompts accurately align with their corresponding positive and negative domain prompt labels, which can be formulated as:
\begin{equation}
\begin{aligned}
\label{eq:DPSPG}
\mathcal{L}_{\text{mse}} &= \mathbb{E}_{d \sim \mathcal{D}_S} \Bigg[ \frac{1}{N_d} \sum_{i=1}^{N_d} \bigg( \left( \hat{\mathbf{v}}^+_i - \mathbf{v}^{+,d} \right)^2 \\
&+ \alpha \cdot \left( \hat{\mathbf{v}}^-_i - \mathbf{v}^{-,d} \right)^2 \bigg) \Bigg],
\end{aligned}
\end{equation}
where $\mathcal{D}_s$ denotes the source domain, and $N_d$ denotes the number of samples in domain $d$. 
The vectors $\mathbf{v}^{+,d}$ and $\mathbf{v}^{-,d}$ represent the obtained positive and negative domain prompt labels, respectively, while $\mathbf{\hat{v}}_i^{+}$ and $\mathbf{\hat{v}}_i^{-}$ denote the corresponding predicted prompt vectors. 
The parameter $\alpha$ balances the contributions of the positive and negative sample errors in the loss function.

\subsubsection{Inference}

As shown in Figure~\ref{fig:intro} (b), during the inference stage, the pre-trained transformer-based prompt generators are used to produce positive and negative domain soft prompts for input images from the target domain. 
The probability that an input image belongs to the \textit{i}-th class is formulated as:
\begin{equation}\begin{aligned}\label{func:inference}
&p(y = y_i \mid \bm{x}) \\
&= \frac
{\exp\left( (\langle t_i^+, \phi(\bm{x}) \rangle - \alpha \cdot \langle t_i^-, \phi(\bm{x}) \rangle) / \tau \right)}
{\sum_{j=1}^K \exp\left( (\langle t_j^+, \phi(\bm{x}) \rangle - \alpha \cdot \langle t_j^-, \phi(\bm{x}) \rangle) / \tau \right)},
\end{aligned}\end{equation}
where $\alpha$ is the balancing hyperparameter, $\tau$ is the temperature parameter, and $K$ denotes the number of classes. 
The positive text feature is computed as $t_i^+ = \psi([G^+(\phi(\bm{x})), v^+, c_i])$, and the negative text feature as $t_i^- = \psi([G^-(\phi(\bm{x})), v^-, c_i])$, 
where $\psi$ denotes the text encoder, and $G^+$ and $G^-$ denote the pre-trained positive and negative transformer-based prompt generators, respectively.

Our DPSPG method directly addresses the problem of prompt variability. 
The positive prompt generator captures the core domain semantics, while the negative prompt generator penalizes deviations from these semantics. 
This dual-path design significantly reduces randomness and ensures that the generated prompts remain tightly clustered. 
As a result, DPSPG achieves highly stable prompt generation and robust generalization to unseen domains.

\subsection{Theoretical Analysis}
\label{subsec:theo}

To further substantiate the effectiveness of our DPSPG method, we provide several theoretical analyses demonstrating how the incorporation of negative prompts improves prompt quality and stabilizes training. 
Specifically, we present proofs based on margin enlargement, gradient norm stabilization, and robustness analysis. 
Together, these results reveal that negative prompt generation contributes to learning more discriminative and stable prompt representations across domains.

\subsubsection{Margin Enhancement}

Let $s_i^+(\mathbf{x}) = \langle t_i^+, \phi(\mathbf{x}) \rangle$ denote the score computed using the positive prompt for class $i$, and similarly, let $s_i^-(\mathbf{x}) = \langle t_i^-, \phi(\mathbf{x}) \rangle$ denote the score computed using the corresponding negative prompt.
When using only positive prompts, the margin between the true class $y$ and any other class $i$ is defined as:
\begin{equation}
\Delta_i^+(\mathbf{x}) = s_y^+(\mathbf{x}) - s_i^+(\mathbf{x}),
\end{equation}
which reflects the model's confidence in preferring the correct class $y$ over the incorrect class $i$ based solely on positive prompt scores.
To further enhance discriminative power, we incorporate both positive and negative prompts. 
In this case, the combined logit for class $i$ is defined as:
\begin{equation}
g_i(\mathbf{x}) = s_i^+(\mathbf{x}) - \alpha\, s_i^-(\mathbf{x}),
\end{equation}
where $\alpha$ is a balancing hyperparameter that controls the contribution of negative prompts. 
Accordingly, the margin between the true class $y$ and an incorrect class $i$ becomes
\begin{equation}
\begin{aligned}
\label{eq:cor_margin}
\Delta_i(\mathbf{x}) &= g_y(\mathbf{x}) - g_i(\mathbf{x}) \\
&= \Delta_i^+(\mathbf{x}) - \alpha \left( s_y^-(\mathbf{x}) - s_i^-(\mathbf{x}) \right),
\end{aligned}
\end{equation}
where $g_y(\mathbf{x})$ and $g_i(\mathbf{x})$ denote the combined logits for the correct class $y$ and the incorrect class $i$, respectively.
By designing the negative prompts such that, for any incorrect class $i \ne y$, the true class $y$ is assigned a lower negative score $s_y^-(\mathbf{x})$ while the incorrect class $i$ is assigned a higher negative score $s_i^-(\mathbf{x})$, we impose the following constraint:
\begin{equation}
s_i^-(\mathbf{x}) \ge s_y^-(\mathbf{x}) + \delta, \quad (\delta > 0),
\end{equation}
where $\delta$ is a positive constant that quantifies the required separation between negative scores.
Substituting this inequality into Equation~\eqref{eq:cor_margin}, we obtain a lower bound on the overall margin:
\begin{equation}
\Delta_i(\mathbf{x}) \ge \Delta_i^+(\mathbf{x}) + \alpha\delta.
\end{equation}

This result shows that incorporating negative prompts effectively enlarges the margin between the correct class and any incorrect class by at least $\alpha \delta$, thereby improving the model's confidence and robustness in classification.

\subsubsection{Gradient Norm Stability and Robustness Analysis}

We analyze how the increased margin, resulting from the incorporation of negative prompts, contributes to more stable gradient behavior during training.
Recall the softmax output $f_i = p(y = i \mid \mathbf{x})$ defined over the logits $g_i(\mathbf{x})$ (Equation~\eqref{func:zsclip}). 
For the cross-entropy loss, the gradient with respect to the logit $g_j$ is given by
\begin{equation}
\label{eq:deriv}
\frac{\partial f_i}{\partial g_j} = \frac{1}{\tau}\, f_i (\delta_{ij} - f_j),
\end{equation}
where $\tau$ is the temperature parameter and $\delta_{ij}$ is the Kronecker delta. 
As the margin $\Delta_i(\mathbf{x}) = g_y(\mathbf{x}) - g_i(\mathbf{x})$ increases, the softmax output $f_y$ for the true class approaches 1, and the corresponding gradient approaches zero. 
This reflects the fact that the sensitivity of the softmax output to logit perturbations diminishes as the margin grows.

We now quantify this behavior. Let $J_f^g(\mathbf{x}) = \frac{\partial f}{\partial g}$ denote the Jacobian of the softmax outputs with respect to the logits. 
In binary classification (between classes $y$ and $i$), Equation~\eqref{eq:deriv} yields:
\begin{equation}
\|J_f^g(\mathbf{x})\| = \left| \frac{\partial f_y}{\partial g_i} \right| = \frac{1}{\tau}\, f_y f_i.
\end{equation}
Using the identity $f_i = \frac{1}{1 + e^{\Delta_i(\mathbf{x}) / \tau}} \le e^{-\Delta_i(\mathbf{x})/\tau}$, we derive the following exponential decay bound:
\begin{equation}
\|J_f^g(\mathbf{x})\| \le \frac{1}{\tau}\, e^{-\Delta_i(\mathbf{x})/\tau}.
\end{equation}

Assuming that the mapping from the input $\mathbf{x}$ to the logits $g$ is $L$-Lipschitz continuous, the chain rule gives:
\begin{equation}
\|J_f(\mathbf{x})\| = \left\| \frac{\partial f}{\partial \mathbf{x}} \right\| \le L\, \|J_f^g(\mathbf{x})\| \le \frac{L}{\tau}\, e^{-\Delta_i(\mathbf{x})/\tau}.
\end{equation}

In our framework, the use of negative prompts increases the margin to at least $\Delta_i^+(\mathbf{x}) + \alpha \delta$, leading to a tighter upper bound on the gradient norm:
\begin{equation}
\|J_f(\mathbf{x})\| \le \frac{L}{\tau}\, e^{-(\Delta_i^+(\mathbf{x}) + \alpha \delta)/\tau}.
\end{equation}
This result shows that negative prompts not only enlarge the margin but also lead to an exponentially smaller gradient norm, thereby promoting smoother optimization and improving robustness.

From a robustness standpoint, the model's response to small input perturbations can be approximated by
\begin{equation}
f(\mathbf{x} + \Delta \mathbf{x}) \approx f(\mathbf{x}) + J_f(\mathbf{x}) \Delta \mathbf{x}.
\end{equation}
A smaller $\|J_f(\mathbf{x})\|$ implies reduced sensitivity to input noise and adversarial perturbations, thereby contributing to greater robustness and better generalization.
In summary, these results establish a direct connection between margin enlargement, achieved through the incorporation of positive and negative prompts, and reduced gradient sensitivity, underscoring the crucial role of negative learning in stabilizing training and enhancing prompt quality within our DPSPG framework.

\section{Experiments}
\label{sec:exp}

\subsection{Experimental Settings}

\subsubsection{Datasets}
We conduct experiments on five benchmark datasets for domain generalization.
PACS~\cite{li2017deeper} consists of four domains, each sharing the same seven categories, with a total of 9,991 images.  
VLCS~\cite{fang2013unbiased} comprises four domains with the same five categories and a total of 10,729 images.  
OfficeHome~\cite{venkateswara2017deep} contains four domains, each consisting of 65 categories related to objects in office and home environments, totaling 15,588 images.  
TerraIncognita~\cite{beery2018recognition} includes 24,778 images of wild animals collected from four distinct regions, covering 10 categories.  
DomainNet~\cite{peng2019moment} is a large-scale dataset comprising six domains and 345 categories, ranging from everyday objects to abstract concepts, with a total of 586,575 images.

\begin{table*}[t]
  \caption{COMPARISONS WITH SOTA METHODS ON PACS AND VLCS FOR MULTI-SOURCE DG IN TERMS OF MEAN LEAVE-ONE-DOMAIN-OUT PERFORMANCE WITH RESNET50 AND VIT-B/16 AS THE BACKBONE. BOLD DENOTES THE BEST SCORES}
  \label{tab:multi-DG_pacs_vlcs}
  \centering
  \resizebox{1.0\textwidth}{!}{
    \begin{tabular}{
      >{\raggedright\arraybackslash}m{2.0cm}|
      *{5}{>{\centering\arraybackslash}m{1.0cm}}|
      *{5}{>{\centering\arraybackslash}m{1.0cm}}}
      \toprule
      \multirow{2}{*}{\textbf{Method}}  & \multicolumn{5}{c|}{\textbf{PACS}}  & \multicolumn{5}{c}{\textbf{VLCS}}  \\
      \cmidrule(lr){2-6} \cmidrule(lr){7-11}
      & Art & Cartoon & Photo & Sketch & Avg 
        & Caltech & LabelMe & Pascal & Sun & Avg \\
      \midrule
      \multicolumn{11}{l}{\textit{ResNet-50 Pre-trained by ImageNet.}} \\
      \midrule
      ERM~\cite{gulrajani2020search} & 84.70 & 80.80 & 97.20 & 79.30 & 85.50 
      & 98.00 & \textbf{64.70} & 75.20 & 71.40 & 77.33 \\
      SWAD~\cite{cha2021swad} & \textbf{89.30} & \textbf{83.40} & \textbf{97.30} & \textbf{82.50} & \textbf{88.13} 
      & \textbf{98.80} & 63.30 & \textbf{79.20} & \textbf{75.30} & \textbf{79.15} \\
      \midrule
      \multicolumn{11}{l}{\textit{ResNet-50 Pre-trained by CLIP.}} \\
      \midrule
      ZS-CLIP~\cite{radford2021learning}   & 90.90 & 93.30 & 99.20 & 79.50 & 90.73
                & 99.43 & 64.87 & 84.13 & 71.55 & 80.00 \\
      LP-CLIP~\cite{radford2021learning}   & 90.77 & 92.67 & 99.10 & 79.80 & 90.58  
                & 99.33 & 61.10 & 81.77 & 76.93 & 79.78 \\
      VP~\cite{bahng2022exploring}        & 90.60 & 92.67 & 99.33 & 78.03 & 90.16  
                & 99.57 & 66.33 & 84.57 & 71.53 & 80.50 \\
      CoOp~\cite{zhou2022learning}      & 92.03 & 93.77 & 98.60 & 80.73 & 91.28  
                & 99.70 & 63.97 & 84.70 & 77.33 & 81.43 \\
      CoCoOp~\cite{zhou2022conditional}    & 93.13 & 94.27 & 99.33 & 80.80 & 91.88  
                & 99.70 & 63.73 & 84.83 & 78.80 & 81.77 \\
      DPL~\cite{zhang2023domain}       & 93.57 & 93.80 & 99.03 & 80.67 & 91.77  
                & \textbf{99.77} & 62.53 & 84.47 & 76.30 & 80.77 \\
      SPG~\cite{bai2024soft}       & 92.77 & 93.83 & 99.47 & 85.13 & 92.80  
                & 99.50 & 68.70 & \textbf{85.37} & \textbf{82.40} & 83.99 \\
      \rowcolor[gray]{0.9} DPSPG (Ours)
                & \textbf{93.95} & \textbf{94.62} & \textbf{99.70} & \textbf{86.37} & \textbf{93.66}  
                & 99.50 & \textbf{71.76} & 84.80 & 80.41 & \textbf{84.12} \\

      \midrule
      \multicolumn{11}{l}{\textit{ViT-B/16 Pre-trained by CLIP.}} \\
      \midrule
      ZS-CLIP~\cite{radford2021learning}   & 97.23 & \textbf{99.07} & \textbf{99.90} & 88.20 & 96.10  
                & 99.93 & 68.63 & 85.87 & 74.77 & 82.30 \\
      LP-CLIP~\cite{radford2021learning}   & 96.17 & 94.73 & 98.70 & 90.07 & 94.92  
                & 95.93 & 63.70 & 76.30 & 74.17 & 77.53 \\
      VP~\cite{bahng2022exploring}        & 96.93 & 98.93 & \textbf{99.90} & 87.27 & 95.76  
                & \textbf{100.00} & 68.47 & \textbf{86.17} & 74.27 & 82.23 \\
      CoOp~\cite{zhou2022learning}      & 97.73 & 98.40 & 99.63 & 90.00 & 96.44  
                & 99.77 & 61.37 & 84.60 & 77.50 & 80.81 \\
      CoCoOp~\cite{zhou2022conditional}    & 97.73 & 98.97 & 99.83 & 90.37 & 96.73  
                & 99.87 & 59.70 & 85.93 & 75.50 & 80.25 \\
      VPT~\cite{jia2022visual}       & 97.90 & 98.90 & \textbf{99.90} & 91.03 & 96.93  
                & 99.87 & 65.47 & 85.53 & 78.47 & 82.33 \\
      DPL~\cite{zhang2023domain}       & 97.77 & 98.50 & \textbf{99.90} & 89.53 & 96.43  
                & 99.83 & 61.47 & 84.57 & 77.83 & 80.93 \\
      MaPLe~\cite{khattak2023maple}     & 97.93 & 98.73 & 99.70 & 89.83 & 96.55  
                & 98.47 & 64.77 & 85.13 & \textbf{81.07} & 82.61 \\
      SPG~\cite{bai2024soft}       & 96.50 & 99.00 & \textbf{99.90} & 91.30 & 96.68  
                & 99.70 & 64.70 & 84.40 & 78.10 & 81.73 \\
      \rowcolor[gray]{0.9} DPSPG (Ours)
                & \textbf{98.05} & 98.59 & 99.64 & \textbf{91.42} & \textbf{96.93}
                & 99.29 & \textbf{69.73} & 83.50 & 78.70 & \textbf{82.81} \\
      \bottomrule
    \end{tabular}
  }
\end{table*}

\begin{table*}[!t]
  \caption{COMPARISONS WITH SOTA METHODS ON OFFICEHOME AND TERRAINCOGNITA FOR MULTI-SOURCE DG IN TERMS OF MEAN LEAVE-ONE-DOMAIN-OUT PERFORMANCE WITH RESNET50 AND VIT-B/16 AS THE BACKBONE. BOLD DENOTES THE BEST SCORES}
  \label{tab:multi-DG_office_terra}
  \centering
  \resizebox{\textwidth}{!}{%
    \begin{tabular}{
      >{\raggedright\arraybackslash}m{2.0cm}|
      *{5}{>{\centering\arraybackslash}m{0.8cm}}|
      *{5}{>{\centering\arraybackslash}m{1.2cm}}}
      \toprule
      \multirow{2}{*}{\textbf{Method}}  & \multicolumn{5}{c|}{\textbf{OfficeHome}}  & \multicolumn{5}{c}{\textbf{TerraIncognita}}  \\
      \cmidrule(lr){2-6} \cmidrule(lr){7-11}
                                        & Art   & Clipart & Product & Real  & Avg   & Location38 & Location43 & Location46 & Location100 & Avg   \\
      \midrule
      \multicolumn{11}{l}{\textit{ResNet-50 Pre-trained by ImageNet.}} \\
      \midrule
      ERM~\cite{gulrajani2020search} & 63.10 & 51.90 & 77.20 & 78.10 & 67.58 
      & 42.50 & 55.60 & 38.80 & 54.30 & 47.80 \\
      SWAD~\cite{cha2021swad} & \textbf{66.10} & \textbf{57.70} & \textbf{78.40} & \textbf{80.20} & \textbf{70.60} 
      & \textbf{44.90} & \textbf{59.70} & \textbf{39.90} & \textbf{55.40} & \textbf{49.98} \\
      \midrule
      \multicolumn{11}{l}{\textit{ResNet-50 Pre-trained by CLIP.}} \\
      \midrule
      ZS-CLIP~\cite{radford2021learning}       & 69.03  & 53.53    & 80.10    & 80.47   & 70.78  & 28.40      & 32.83      & 23.97      & 10.13        & 23.83  \\
      LP-CLIP~\cite{radford2021learning}       & 61.97  & 48.97    & 73.60    & 77.43   & 65.49  & 32.97      & 42.73      & 31.87      & 24.40        & 32.99  \\
      VP~\cite{bahng2022exploring}            & 67.67  & 52.53    & 80.03    & 80.40   & 70.16  & 28.77      & 33.97      & 26.83      & 12.60        & 25.54  \\
      CoOp~\cite{zhou2022learning}          & 71.27  & 57.07    & 83.20    & 83.53   & 73.77  & 25.60      & 43.50      & 34.50      & 29.23        & 33.21  \\
      CoCoOp~\cite{zhou2022conditional}        & 71.33  & 56.73    & 83.77    & 83.33   & 73.79  & 35.90      & 42.10      & 32.50      & 25.80        & 34.08  \\
      DPL~\cite{zhang2023domain}           & 71.50  & 56.33    & 84.03    & 83.13   & 73.75  & 36.03      & 41.07      & 32.90      & 27.60        & 34.40  \\
      SPG~\cite{bai2024soft}           & 71.30  & 55.60    & 84.80    & \textbf{83.40}   & 73.78  & 45.77      & 38.90      & 32.10      & 36.80        & 38.39  \\
      \rowcolor[gray]{0.9} DPSPG (Ours)  & \textbf{71.69}  & \textbf{57.62}    & \textbf{85.47}    & 82.60   & \textbf{74.35}  & \textbf{48.57}      & \textbf{45.06}      & \textbf{36.70}      & \textbf{54.80}        & \textbf{46.28}  \\
      \midrule
      \multicolumn{11}{l}{\textit{ViT-B/16 Pre-trained by CLIP.}} \\
      \midrule
      ZS-CLIP~\cite{radford2021learning}       & 80.13  & 70.03    & 88.17    & 88.97   & 81.83  & 20.50      & 32.80      & 29.63      & 52.37       & 33.83  \\
      LP-CLIP~\cite{radford2021learning}       & 73.53  & 69.90    & 87.37    & 86.43   & 79.31  & 48.00      & 50.50      & 43.80      & 44.00        & 46.58  \\
      VP~\cite{bahng2022exploring}            & 79.80  & 69.10    & 87.43    & 88.57   & 81.23  & 20.23      & 34.27      & 32.80      & 52.30        & 34.90  \\
      CoOp~\cite{zhou2022learning}          & 81.23  & 71.97    & 89.70    & 89.20   & 83.02  & 54.83      & 47.37      & 41.13      & 45.47        & 47.20  \\
      CoCoOp~\cite{zhou2022conditional}        & 81.80  & 71.73    & 90.33    & 89.87   & 83.40  & 51.63      & 46.90      & 39.30      & 43.17        & 45.25  \\
      VPT~\cite{jia2022visual}           & 80.93  & 72.50    & 90.03    & 89.37   & 83.21  & 46.77      & 52.80      & 41.83      & 45.50        & 46.73  \\
      DPL~\cite{zhang2023domain}           & 81.03  & 71.37    & 91.10    & 89.6   & 83.28  & 54.33      & 48.97      & 41.63      & 41.60        & 46.63  \\
      MaPLe~\cite{khattak2023maple}         & 81.63  & 72.63    & 90.23    & 89.53   & 83.50  & 52.43      & \textbf{53.00}      & 44.10      & 56.30        & 51.46  \\
      SPG~\cite{bai2024soft}           & 81.60  & 72.70    & 90.20    & \textbf{89.90}   & 83.60  & 51.00      & 49.20      & \textbf{50.70}      & 49.80        & 50.18  \\
      \rowcolor[gray]{0.9} DPSPG (Ours)  & \textbf{82.41}  & \textbf{73.63}    & \textbf{91.03}    & 89.58   & \textbf{84.16}  & \textbf{56.05}      & 50.65      & 42.43      & \textbf{61.04}        & \textbf{52.54}  \\
      \bottomrule
    \end{tabular}
  }
\end{table*}

\begin{table*}[t]
\caption{COMPARISONS WITH SOTA METHODS ON DOMAINNET FOR MULTI-SOURCE DG IN TERMS OF MEAN LEAVE-ONE-DOMAIN-OUT PERFORMANCE WITH RESNET50 AND VIT-B/16 AS THE BACKBONE. BOLD DENOTES THE BEST SCORES}
\label{tab:multi-DG_domainnet}
\centering
\begin{adjustbox}{width=1.0\textwidth,height=6.2cm, keepaspectratio}
\begin{tabular}{
>{\raggedright\arraybackslash}m{2.0cm}|
>{\centering\arraybackslash}m{1.5cm}
>{\centering\arraybackslash}m{1.5cm}
>{\centering\arraybackslash}m{1.5cm}
>{\centering\arraybackslash}m{1.5cm}
>{\centering\arraybackslash}m{1.5cm}
>{\centering\arraybackslash}m{1.5cm}
>{\centering\arraybackslash}m{1.5cm}
}
\toprule
Method & Clipart & Infograph & Painting & Quickdraw & Real & Sketch & Avg \\
\midrule
\multicolumn{6}{l}{\textit{ResNet-50 Pre-trained by ImageNet.}} \\
\midrule
ERM~\cite{gulrajani2020search} & 63.00 & 21.20 & 50.10 & 13.90 & 63.70 & 52.00 & 43.98 \\
SWAD~\cite{cha2021swad} & \textbf{66.00} & \textbf{22.40} & \textbf{53.50} & \textbf{16.10} & \textbf{65.80} & \textbf{55.50} & \textbf{46.55} \\
\midrule
\multicolumn{6}{l}{\textit{ResNet-50 Pre-trained by CLIP.}} \\
\midrule
ZS-CLIP~\cite{radford2021learning} & 52.73 & 40.53 & 53.20 & 5.70 & 77.07 & 49.27 & 46.42 \\
LP-CLIP~\cite{radford2021learning} & 34.60 & 24.73 & 35.33 & 4.13 & 28.17 & 35.87 & 27.14 \\
VP~\cite{bahng2022exploring} & 52.43 & 40.33 & 52.70 & 5.27 & 76.83 & 47.13 & 45.78 \\
CoOp~\cite{zhou2022learning} & 57.03 & 43.93 & 58.13 & 7.80 & 78.77 & 52.57 & 49.71 \\
CoCoOp~\cite{zhou2022conditional} & 57.00 & 44.00 & 58.33 & 7.83 & 78.90 & 52.00 & 49.68 \\
DPL~\cite{zhang2023domain} & 56.73 & 43.90 & 57.90 & 7.90 & 78.17 & 53.03 & 49.61 \\
SPG~\cite{bai2024soft} & \textbf{57.30} & 41.70 & \textbf{58.30} & 7.90 & 79.03 & 55.23 & 49.91 \\
\rowcolor[gray]{0.9} DPSPG (Ours) & 55.65 & \textbf{45.04} & 57.30 & \textbf{8.10} & \textbf{79.62} & \textbf{55.30} & \textbf{50.17} \\
\midrule
\multicolumn{6}{l}{\textit{ViT-B/16 Pre-trained by CLIP.}} \\
\midrule
ZS-CLIP~\cite{radford2021learning} & 70.20 & 46.27 & 65.03 & 13.00 & 83.00 & 62.03 & 56.59 \\
LP-CLIP~\cite{radford2021learning} & 62.90 & 35.37 & 56.77 & 11.33 & 65.77 & 56.73 & 48.14 \\
VP~\cite{bahng2022exploring} & 70.10 & 45.50 & 64.57 & 14.07 & 82.73 & 61.97 & 56.49 \\
CoOp~\cite{zhou2022learning} & 72.70 & 50.20 & 68.50 & 15.63 & 84.23 & 65.93 & 59.53 \\
CoCoOp~\cite{zhou2022conditional} & 72.13 & 50.37 & 67.90 & 15.83 & 84.37 & 65.53 & 59.36 \\
VPT~\cite{jia2022visual} & 71.03 & 48.50 & 66.17 & 16.27 & 83.63 & 65.23 & 58.47 \\
DPL~\cite{zhang2023domain} & 72.47 & 50.40 & 68.30 & 15.83 & 83.90 & 66.00 & 59.48 \\
MaPLe~\cite{khattak2023maple} & \textbf{73.53} & 50.70 & 67.60 & 17.00 & 83.43 & 66.30 & 59.76 \\
SPG~\cite{bai2024soft} & 68.70 & 50.20 & \textbf{73.20} & 16.06 & 83.33 & \textbf{68.47} & 59.99 \\
\rowcolor[gray]{0.9} DPSPG (Ours) & 72.54 & \textbf{50.74} & 69.70 & \textbf{17.22} & \textbf{84.50} & 67.40 & \textbf{60.35} \\
\bottomrule
\end{tabular}
\end{adjustbox}
\end{table*}

\subsubsection{Baselines} 
We evaluate our method against two categories of baselines.
For traditional DG methods, we report results for ERM~\cite{gulrajani2020search} and SWAD~\cite{cha2021swad}.
For CLIP-based methods, we consider several variants. 
Specifically, we compare against zero-shot CLIP (ZS-CLIP)~\cite{radford2021learning}, which uses handcrafted prompts such as \texttt{`a photo of a \{class\}'}, and linear probing of CLIP (LP-CLIP), where a linear classification head is trained while keeping the CLIP visual encoder frozen. 
We also compare with prompt tuning methods, which are further divided into two subgroups: fixed prompt learning methods, including VP~\cite{bahng2022exploring}, CoOp~\cite{zhou2022learning}, VPT~\cite{jia2022visual}, and MaPLe~\cite{khattak2023maple}; and dynamic prompt learning methods, including CoCoOp~\cite{zhou2022conditional}, DPL~\cite{zhang2023domain}, and SPG~\cite{bai2024soft}.

\subsubsection{Implementation Details} 
In the dual-path domain prompt label learning phase of DPSPG, we learn domain-specific positive and negative soft prompts $\mathbf{v}^+$ and $\mathbf{v}^-$ for each domain across the five datasets. 
The positive prompts are initialized with the context phrase ``a photo of a,'' and the negative prompts with ``a photo without a,'' both with a context length of 4.
We optimize the learnable text vectors following the CoOp framework~\cite{zhou2022learning} using stochastic gradient descent (SGD) with an initial learning rate of 2e-3.
Training is performed for 70 epochs with a cosine annealing learning rate schedule. 
The batch size is set to 32, except for DomainNet, where it is reduced to 8 due to the dataset's larger size. 
The positive and negative soft prompts achieving the best validation performance are selected as the final domain prompt labels.

In the prompt generator pre-training phase, we train the transformer-based prompt generators on each domain. 
We use the AdamW optimizer with a weight decay of 1e-3 and beta values of (0.9, 0.999). 
The initial learning rate is set between 2e-5 and 2e-3 depending on the dataset, with a linear warm-up phase at 1e-5 for the first four epochs.
Training is conducted for 50 epochs with a cosine annealing schedule. 
During evaluation, we set the negative prompt loss weight $\alpha$ to 0.2 and apply early stopping based on validation performance for certain domains.

\subsection{Main Results}

We follow the leave-one-domain-out evaluation protocol~\cite{gulrajani2020search} for multi-source domain generalization. 
In this protocol, one domain is excluded from the training set in each round, and the model is tested on this excluded domain. This process is repeated iteratively until each domain has been held out and tested.

We compare DPSPG against a wide range of state-of-the-art (SOTA) methods on the PACS, VLCS, OfficeHome, TerraIncognita, and DomainNet datasets using ResNet-50 and ViT-B/16 backbones, as summarized in Tables~\ref{tab:multi-DG_pacs_vlcs}, \ref{tab:multi-DG_office_terra}, and \ref{tab:multi-DG_domainnet}. 
We report the per-domain performance as well as the three-run average leave-one-domain-out accuracy of all baseline methods and our DPSPG. 
Our results consistently demonstrate that DPSPG achieves superior performance over all other state-of-the-art methods.

\begin{table*}[tbp]
\caption{COMPARISONS WITH DIFFERENT COMPONENTS OF ABLATION ON DOMAIN GENERALIZATION BENCHMARK PACS DATASET FOR MULTI-SOURCE DG PERFORMANCE WITH RESNET50 AS THE BACKBONE. POS DENOTES INCORPORATING THE POSITIVE PROMPT, NEG DENOTES INCORPORATING THE NEGATIVE PROMPT. TRANS DENOTES THE TRANSFORMER MODEL. BOLD DENOTES THE BEST SCORES}
\label{tab:ablation}
\centering
\resizebox{0.9\textwidth}{!}{
\begin{tabular}{
>{\centering\arraybackslash}m{0.7cm}
>{\centering\arraybackslash}m{0.6cm}
>{\centering\arraybackslash}m{0.6cm}
>{\centering\arraybackslash}m{1cm}
>{\centering\arraybackslash}m{1cm} |
>{\centering\arraybackslash}m{0.8cm}
>{\centering\arraybackslash}m{1.3cm}
>{\centering\arraybackslash}m{1.4cm}
>{\centering\arraybackslash}m{1.1cm}
>{\centering\arraybackslash}m{0.9cm}}
\toprule
\multirow{2}{*}{Exp} & \multicolumn{4}{c|}{Component Ablation} &  &  &  &  &  \\
\cmidrule(r){2-5}
& Pos & Neg & CGAN & Trans & Art & Cartoon & Painting & Sketch & Avg \\

\midrule
\#1  & $\checkmark$  &  & - & - & 92.00 & 93.81 & 98.56 & 80.70 & 91.27 \\
\#2  & $\checkmark$  & $\checkmark$ & $\checkmark$ & - & 92.72 & 90.53 & 99.04 & 83.63 & 91.48 \\
\#3 & $\checkmark$  &  & - & $\checkmark$ & 93.36 & 93.52 & 99.40 & 85.77 & 93.01 \\
\#4  & $\checkmark$ & $\checkmark$  & - & $\checkmark$ & \textbf{93.95} & \textbf{94.62} & \textbf{99.70} & \textbf{86.37} & \textbf{93.66} \\
\bottomrule
\end{tabular}
}
\end{table*}

\textbf{Results on PACS.} 
Specifically, on the PACS dataset using the ResNet-50 backbone, DPSPG achieves an average accuracy of 93.66\%, representing a 0.86\% improvement over SPG. 
Moreover, DPSPG attains the highest accuracy across all four domains, including 86.37\% on the Sketch domain, which is widely regarded as the most challenging domain in PACS.

\textbf{Results on VLCS.} 
On the VLCS dataset, DPSPG achieves state-of-the-art accuracies of 84.12\% with ResNet-50 and 82.81\% with ViT-B/16. 
Notably, DPSPG demonstrates substantial improvements on LabelMe, the most challenging domain in VLCS, surpassing previous methods by 3.06\% with ResNet-50 and 1.10\% with ViT-B/16, where fixed prompt learning approaches often struggle.

\textbf{Results on OfficeHome.} 
On the OfficeHome dataset, DPSPG consistently outperforms all other methods across both backbones. 
It achieves 74.35\% accuracy with ResNet-50 and 84.16\% with ViT-B/16, setting new state-of-the-art results in three out of the four domains for each backbone.

\textbf{Results on TerraIncognita.} 
On the TerraIncognita dataset with the ResNet-50 backbone, DPSPG surpasses the previous best method, SPG, by 7.89\%, establishing new state-of-the-art results across all four domains. 
These results underscore the pivotal role of negative learning in enhancing model robustness and promoting generalization under distribution shifts.

\textbf{Results on DomainNet.} 
On the DomainNet dataset, DPSPG achieves state-of-the-art results, attaining 50.17\% accuracy with ResNet-50 and 60.35\% with ViT-B/16. 
Using ResNet-50 as the backbone, DPSPG achieves particularly strong performance with 45.04\% accuracy on Infograph and an 8.10\% improvement on Quickdraw compared to other methods, two of the most challenging domains in DomainNet. 
These results underscore DPSPG's ability to stabilize prompt generation under severe domain shifts and effectively address prompt variability in the most difficult transfer scenarios.

Overall, DPSPG achieves an average improvement of approximately 3.94\% across all five benchmarks using the ResNet-50 backbone, establishing a new state-of-the-art for the multi-source domain generalization task. 
These results highlight the potential of the dual-path stable prompt generation paradigm in enhancing the generalization capability of prompt learning.

\subsection{Visualization}

\begin{figure}[tbp]
  \centering
  \includegraphics[width=0.5\textwidth]{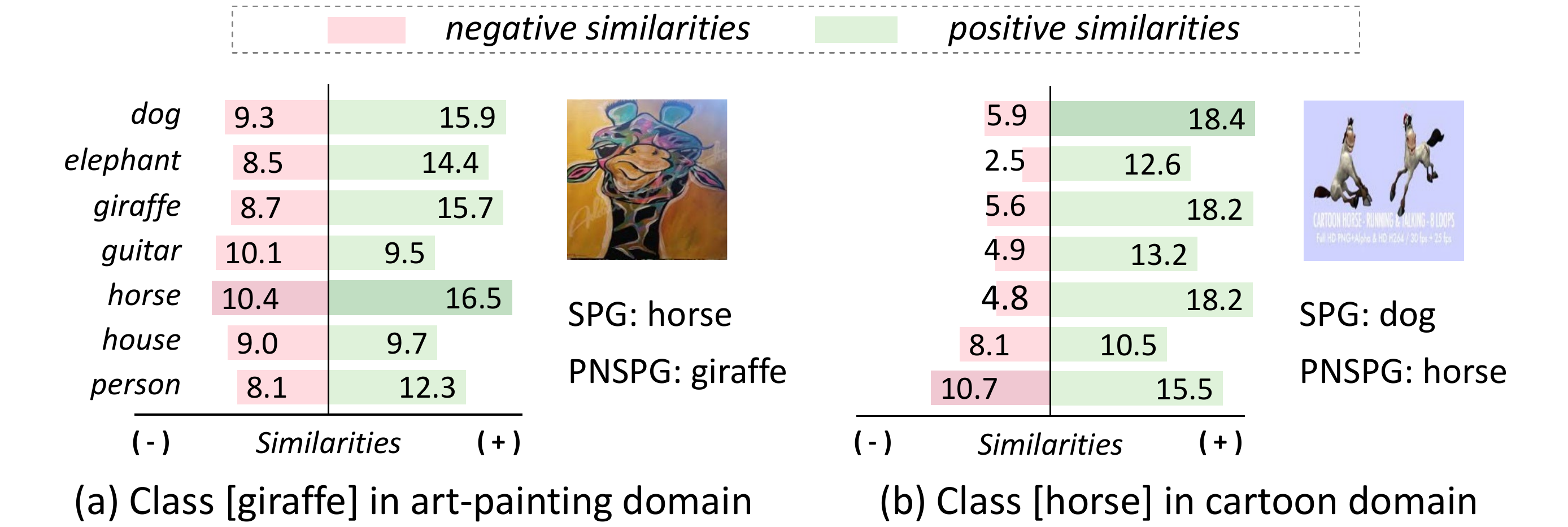}
  \caption{Two examples during inference. Compared with SPG, DPSPG enhances its predictive capabilities by incorporating negative learning.}
  \label{fig:visual}
\end{figure}

As illustrated in Figure~\ref{fig:visual}, we present two example images from the PACS dataset. 
Positive and negative similarity scores are computed by comparing the text features generated from positive and negative soft prompts with the visual features of the query images, respectively. 
In some cases, the positive prompt alone results in a weak or ambiguous match to the correct class. 
The negative prompt, however, effectively suppresses similarities to incorrect classes, thereby sharpening and stabilizing the positive score. 
This complementary interaction, where negative signals suppress noise and positive signals reinforce the correct information, leads to more accurate and consistent prompt representations compared to using positive prompts alone. 
These results highlight the effectiveness of our DPSPG method, demonstrating that generating both positive and negative soft prompts enhances generalization performance in prompt learning, and provides distinct advantages in improving model robustness and accuracy.

\subsection{Ablation and Analysis Study}

\subsubsection{Ablation on the Negative Prompts}

As illustrated in Table~\ref{tab:ablation}, we present the comparison across four different conditions: (1) Positive prompt learning only, equivalent to CoOp~\cite{zhou2022learning}; (2) SPG~\cite{bai2024soft} integrated with negative learning; (3) Our DPSPG method without incorporating negative learning; and (4) Our complete DPSPG model with all components included.

Removing negative prompts (\#3 vs.\ \#4) consistently leads to performance degradation, highlighting the importance of negative learning. 
The inclusion of negative prompts sharpens the prompt space by suppressing irrelevant correlations, resulting in more stable and transferable representations across domains. 
By jointly leveraging positive and negative prompts, DPSPG learns clearer and more robust prompt representations, thereby improving domain generalization.

\begin{figure*}[htbp]
  \centering
  \includegraphics[width=0.9\textwidth]{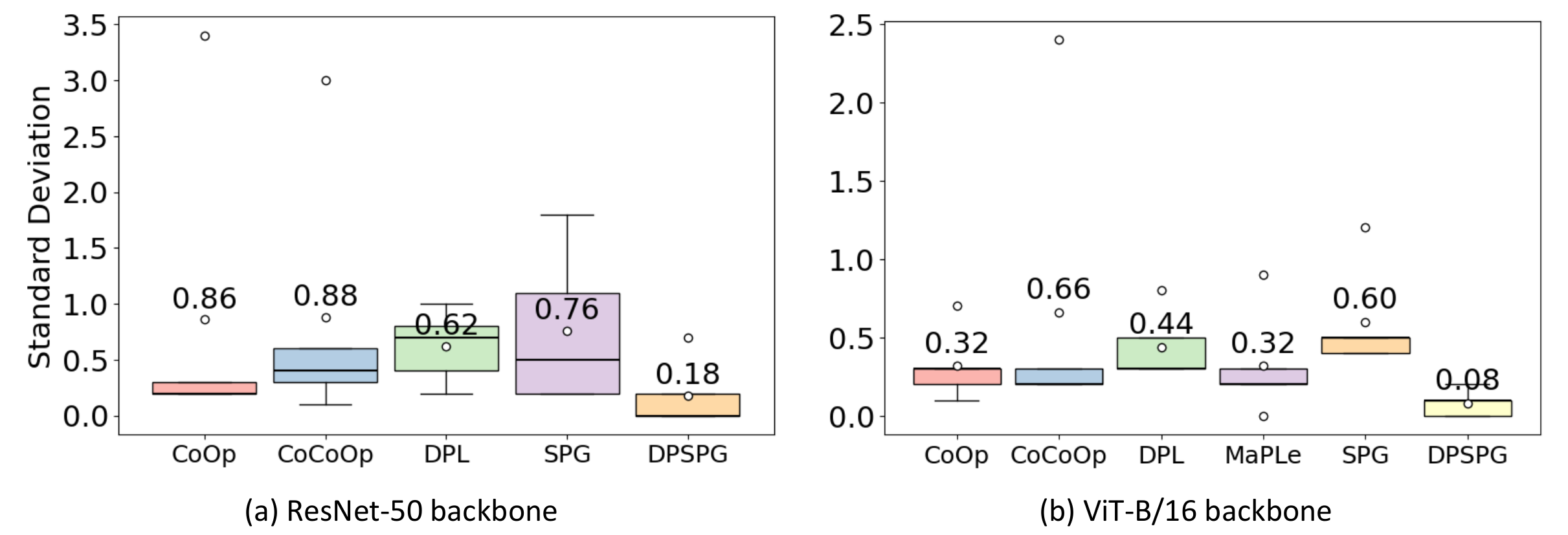}
  \caption{Standard deviation of leave-one-domain-out accuracies across five datasets for various CLIP-based prompt learning methods using (a) ResNet-50 and (b) ViT-B/16 backbones. DPSPG consistently exhibits the lowest standard deviation across domains, which has the narrowest interquartile ranges and shortest whiskers, indicating greater generalization stability and robustness of its dual-path prompt generation strategy.
  }
  \label{fig:std_box}
\end{figure*}

\subsubsection{Ablation on the Prompt Generator}
As shown in Table~\ref{tab:ablation}, replacing our transformer-based generator with a CGAN (\#2 vs.\ \#4) leads to a drop in accuracy. 
The transformer backbone better captures long-range dependencies and domain-specific nuances, resulting in more reliable soft prompts. 
In contrast, the CGAN exhibits greater instability and sensitivity to hyperparameters, producing noisier prompts and weaker generalization, thereby underscoring the importance of a stable transformer architecture for robust prompt generation.

\subsubsection{Sensitivity Analysis of Parameter}
\begin{figure}[tbp]
  \centering
  \includegraphics[width=0.46\textwidth]{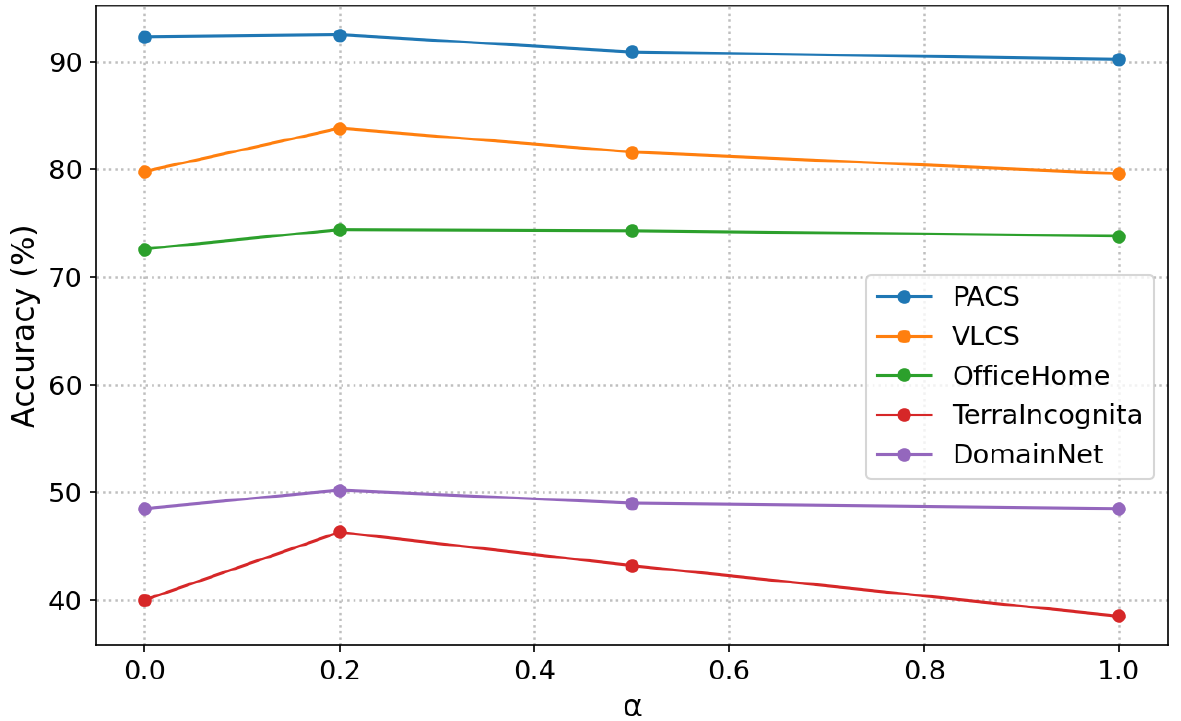}
  \caption{Sensitivity analysis of parameter $\alpha$ on five DG benchmark datasets for multi-source DG performance with ResNet50 as the backbone.}
  \label{fig:abla_alpha}
\end{figure}

As shown in Figure~\ref{fig:abla_alpha}, we evaluate the impact of the combination weight $\alpha$ used in Equation~\ref{eq:DPSPG}. 
Across all five datasets, a consistent trend emerges: moderate values of $\alpha$, particularly around 0.2, achieve the best performance, while the overall variation remains small across a broad range. 
This indicates that DPSPG is generally robust to the choice of $\alpha$ and maintains strong performance without requiring precise hyperparameter tuning. 
Notably, DomainNet exhibits slightly higher sensitivity to $\alpha$, with larger performance fluctuations compared to other datasets, highlighting the potential benefits of careful parameter selection.

\begin{table}[tbp]
\small
\centering
\caption{EFFICIENCY AND TRAINING STABILITY COMPARISON WITH SPG ON TERRAINCOGNITA DATASET. TIME DENOTES THE TRAINING TIME OF THE PROMPT GENERATOR, STD. DENOTES THE STANDARD DEVIATION OF ACCURACY OF THE LAST 10 EPOCHS}
\label{tab:efficiency}
\resizebox{0.96\linewidth}{!}{
\begin{tabular}{lcccccc}
\toprule
Method & Time & Epochs & GFLOPs & Param. & Std. & Acc.   \\ 
\midrule
SPG & 12 hour & \textbf{50} & 0.227 & \textbf{5.0M} & 9.8 & 38.39 \\
DPSPG & \textbf{2 hour} & \textbf{50} & \textbf{0.126} & 18.9M & \textbf{3.7} & \textbf{46.28} \\
\bottomrule
\end{tabular}
}
\end{table}

\subsubsection{Comparison of Efficiency}
We compare the efficiency of SPG and DPSPG in Table~\ref{tab:efficiency}. 
Although DPSPG has a slightly higher parameter count, it remains highly manageable. 
Notably, DPSPG is six times faster and requires half the FLOPs compared to SPG. 
This improvement is largely attributed to the CGAN architecture used in SPG, which involves both a generator and a discriminator, resulting in doubled gradient propagation overhead. 
In contrast, DPSPG achieves higher accuracy with superior computational efficiency, demonstrating its effectiveness in balancing model complexity and performance.

\subsubsection{Comparison of Training and Inference Stability}
As shown in the last column of Table~\ref{tab:efficiency}, DPSPG exhibits a significantly lower standard deviation compared to SPG, which often suffers from high variability, demonstrating more stable training process. 
This indicates that DPSPG not only achieves superior average performance but also delivers more consistent results, enhancing its reliability for practical applications.

Moreover, we visualize the performance distribution as box plots over the five datasets in Figure~\ref{fig:std_box}. 
It is evident that DPSPG consistently produces the narrowest boxes and shortest whiskers among all CLIP-based methods, indicating more stable inference performance. 
This reduction in variance demonstrates that, beyond improving mean accuracy, DPSPG markedly stabilizes prompt generation across domains, further validating the effectiveness of the dual-path mechanism in domain generalization.

\section{Conclusion}
\label{sec:conclu}

In this work, we propose Dual-Path Stable Soft Prompt Generation (DPSPG), a novel prompt learning framework designed to address the prompt variability problem in domain generalization. 
By introducing a dual-path mechanism that generates both positive and negative prompts, DPSPG explicitly enhances the stability and consistency of prompt generation across domains. 
Through theoretical analysis, we demonstrate that the incorporation of negative prompts enlarges the effective margin, resulting in smaller gradient norms and improved robustness to input perturbations. 
This margin-based stability ensures smoother optimization dynamics and better generalization to unseen domains. 
Extensive experiments on five domain generalization benchmarks validate the effectiveness of our method, with DPSPG consistently outperforming existing state-of-the-art methods.

\section*{Acknowledgement}
This work was supported by the National Natural Science Foundation of China with grant numbers (U21A20485, 62436005, and 62088102), and the Fundamental Research Funds for Xi’an Jiaotong University under Grants xzy022024012.

\bibliographystyle{IEEEtran}
\bibliography{refs}

\end{document}